\documentclass[conference]{IEEEtran}

\usepackage{amssymb,amsmath}
\usepackage{multirow}
\usepackage[utf8]{inputenc}
\usepackage[vietnam,english]{babel}
\usepackage{tabularx}
\ifCLASSINFOpdf
 \usepackage[pdftex]{graphicx}
\else
 \usepackage[dvips]{graphicx}
\fi
\begin{document}
\selectlanguage{english}
%
\title{Error Analysis for Vietnamese Dependency Parsing}


\author{\IEEEauthorblockN{Kiet Van Nguyen}
\IEEEauthorblockA{Department of Information Science and Engineering\\
University of Information Technology\\Vietnam National University --
Ho Chi Minh City, Vietnam\\
Email: kietnv@uit.edu.vn}
\and
\IEEEauthorblockN{Ngan  Luu-Thuy Nguyen}
\IEEEauthorblockA{Faculty of Computer Science\\University of Information Technology\\Vietnam National University -- 
Ho Chi Minh City, Vietnam\\
Email: ngannlt@uit.edu.vn}}


%
%
%
%
\maketitle
\begin{abstract}
Dependency parsing is needed in different applications of natural language processing. In this paper, we present a thorough error analysis for dependency parsing for the Vietnamese language, using two state-of-the-art parsers:  MSTParser and MaltParser. The error analysis results provide us insights in order to improve the performance of dependency parsing for the Vietnamese language.
\end{abstract}

\IEEEpeerreviewmaketitle
\section{Introduction}

Dependency parsing is one of the fundamental problems in natural language processing. In dependency parsing, there are two main approaches: grammar-based and data-driven. Recent works on dependency parsing mainly focus on data-driven parsing models.\\
\indent The problem of dependency parsing is described as follows:

\indent \textbf{Input}: A sentence S consists of n words: S = $w_0$, $ w_1$, $w_2$, ..., $w_n$, where $w_0$ = ROOT.\\
\indent \textbf{Output}: A connected, acyclic, and single-head dependency graph G (V, A), in which:
\begin{itemize}
	\item V = \{\textit{0, 1, ..., n}\} is the vertex set;
	\item A is the arc set, i.e, \textit{(i, j, $l_k$)} $\in$ A represents a dependency (arc) from $w_i$ to $w_j$ with label $l_k$ $\in$ L;
	\item L = \{$l_1$, $l_2$, …, $l_L$\} is a set of permissible arc labels.
\end{itemize}

Dependency tree structures can be divided into two types: projective and non-projective. A non-projective dependency tree contains crossing arcs, while projective dependency trees do not.

\begin{figure}[b]
\selectlanguage{english}
\centering
\includegraphics[scale=0.5]{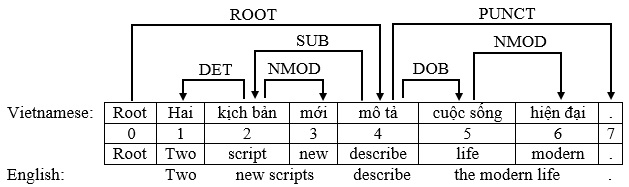}
\caption{Dependency graph for a Vietnamese sentence.}
\label{fig:sentence_1}
\end{figure}
\selectlanguage{vietnam}

Figure \ref{fig:sentence_1} shows an example of parsing results for a Vietnamese sentence \textit{Hai kịch bản mới mô tả cuộc sống hiện đại (English translation: Two new scripts describe the mordern life.)}. 
We can see multiple instances of labeled dependency relations such as the one from the verb \textit{mô tả (describe)} to \textit{kịch bản (scripts)} with the SUB label indicating that \textit{kịch bản (scripts)} is the head of the syntactic subject of the verb.

Dependency parsing has gained a wide interest in the research community of natural language processing during the past decade. Dependency parsing has been successfully employed for many applications such as information retrieval, text summary, machine translation, and  question answering. Large and prestigious conferences in the field, including ACL, EACL, and COLING, have constantly provided tutorials on dependency parsing \cite{Tutorial_2006, Tutorial_2010, Tutorial_2014_2, Tutorial_2014_1, Tutorial_2013}. In particular, the 2006 and 2007 CoNLL Shared Tasks \cite{CoNLL_2006, CoNLL_2007} led to a boom in study on data-driven dependency parsing on many languages: from 13 languages and 19 systems (CoNLL, 2006) to 19 languages and 23 systems (CoNLL, 2007). 

State-of-the-art methods on dependency parsing for the Vietnamese language achieved only less than 80\% \cite{DP_Vi_2, DP_Vi_1}.
To understand why parsing performance is low and know how to improve it, we carried out a thorough error analysis based on two state-of-the-art data-driven dependency parsing systems, MSTParser and MaltParser. 

%
%
%
\section{Related Works}
%
%
%
%

%
%
%
%
\subsection{State-of-the-art Dependency Parsing}
\begin{figure}[t!]
\selectlanguage{english}
\centering
\includegraphics[scale=0.38]{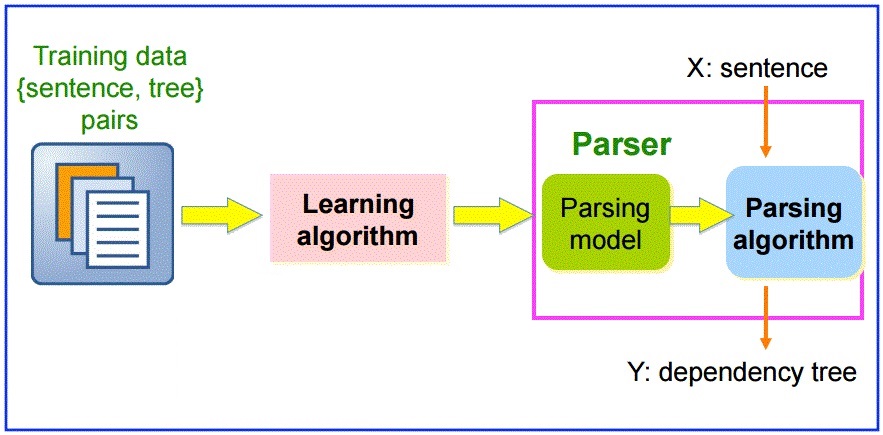}
\caption{A data-driven dependency parsing model \cite{Tutorial_2010}.}
\label{fig:dpmodel}
\end{figure}

A typical data-driven model for dependency parsing is shown in Figure \ref{fig:dpmodel} which is borrowed from \cite{Tutorial_2010}. It includes three important components: learning algorithm, parsing model, and parsing algorithm. Depending on the parsing algorithm, data-driven dependency parsing models can be divided two types: graph-based and transition-based. The combination of graph-based models and transition-based models forms hybrid models. In this work, we focus on the first two approaches: graph-based and transition-based.

Graph-based dependency parsing systems \cite{Error_2} parameterize models over dependency subgraphs and learn these parameters to score correct graphs above incorrect ones for every sentence in a training set. The parsing systems try to search the highest-scoring dependency graph among the set of all possible dependency graphs, which can be considered as a global inference process. MSTParser is a graph-based dependency parser developed by the group of McDonald et al. since 2006. This system is freely available for research purpose\footnotemark[1]. 

As described in paper \cite{Error_2}, transition-based dependency parsing systems  parameterize models over transitions from a state to other in an abstract state machine. Parameters in these models are learned using classification techniques to predict the most probable transition out of a set of possible transitions given a state history. 
MaltParser is a transition-based dependency parser which was developed by Nirve et al. This parser is freely available for research purpose\footnotemark[2]. A comparison of the characteristics of MSTParser and MaltParser is briefly shown in Table \ref{table:table_1}. 

\begin{table}[h]
\selectlanguage{english}
\caption{Comparing between two data-driven parsing systems }
\label{table:table_1}
\begin{tabular}{|l|l|l|}
\hline
\multicolumn{1}{|c|}{\textbf{Characteristic}} & \multicolumn{1}{c|}{\textbf{MST\footnotemark[3]}} & \multicolumn{1}{c|}{\textbf{Malt}} \\ \hline
Inference                                     & Exhaustive                        & Greedy                             \\ \hline
Training                                      & Global structure learning         & Local decision learning            \\ \hline
Features                                      & Local features                    & Rich decision history              \\ \hline
Fundamental trade-off                         & Golbal learning and inference     & Rich feature space                 \\ \hline
\end{tabular}
\end{table}

\footnotetext[1]{http://www.seas.upenn.edu/~strctlrn/MSTParser/MSTParser.html}
\footnotetext[2]{http://www.maltparser.org/}
\footnotetext[3]{In all tables, MSTParser and MaltParser are referred to as MST and Malt, respectively, for short.}

Two commonly-used evaluation metrics for dependency parsing are the unlabeled attachment score (UAS) and the labeled attachment score (LAS) \cite{CoNLL_2007}.  UAS is the percentage of tokens that are correctly assigned to the heads. LAS is the percentage of tokens for which a system has predicted the correct head and dependency type.

\begin{table}[t!]
\selectlanguage{english}
\caption{Labeled parsing accuracy for top scoring systems at CoNLL-X}
\label{table:performance}
\centering
\begin{tabular}{|l|l|l|l|l|l|}
\hline
\textbf{Language} & \textbf{MST} & \textbf{Malt} & \textbf{Language} & \textbf{MST} & \textbf{Malt} \\ \hline
                  Arabic      &66.91       &66.71         &Japanese         &90.71          &91.65                  \\ \hline
Bulgarian      &87.57       &87.41         &Portuguese         &86.82          &87.60                  \\ \hline
Chinese      &85.90       &86.92         &Slovene         &73.44          &70.30                  \\ \hline
Czech      &80.18       &78.42         &Spanish         &82.25          &81.29                  \\ \hline
Danish      &84.79       &84.77         &Swedish         &82.55          &84.58                  \\ \hline
Dutch      &79.19       &78.59         &Turkish         &63.19          &65.68                  \\ \hline
German      &87.34       &85.82         &Overall         &80.83          &80.75                  \\ \hline
\end{tabular}
\end{table}

Table \ref{table:performance} shows the results of the two top performing systems in the CoNLL-X shared task, developed by two groups McDonald et al. (2006) and Nirve et al.(2006). This table is borrowed from \cite{Tutorial_2014_1} for the sake of comparison with the parsing results of the Vietnamese language. 
For all 13 languages,  MSTParser and MaltParser achieved in average 80.83\% and 80.75\% in accuracy respectively, as shown in Table \ref{table:performance}, which are not significantly differ ($\Delta$ = 0.08\%). 
The works \cite{Error_2, Error_1} characterize the difference in errors made by a global, exhaustive, graph-based parsing system (MSTParser) and a local, greedy, transition-based parsing system (MaltParser). 


\subsection{Vietnamese Dependency Treebank}
The Vietnamese Treebank \cite{VietTreebank} was developed as part of the national project -- Vietnamese Language and Speech Processing. This treebank contains about 10.200 phrase-structure trees (about 220.000 tokens). Vietnamese Dependency Treebank (VnDT) contains dependency structures transferred from Vietnamese Treebank following Dat Nguyen's approach \cite{DP_Vi_2}. The VnDT treebank contains 33 dependency types. Table \ref{table:DPVnDT} shows the distributions of all the dependency types. The proportion of non-projective structures in VnDT is 4.49\%. The percentage of sentences with length less than 30 tokens is 80\%. The percentage of sentences with length over 20 tokens accounts for 45.61\%. It can be seen that the average length of the sentences in VnDT is 21.45 tokens, which is longer than most of other languages. In 13 languages in Table  \ref{table:performance}, there are only three languages (Arabic, Portuguese, and Spanish) with average sentence length over 20 tokens \cite{Error_2}.

\begin{table}[h]
\selectlanguage{english}
\caption{Percentage of dependency types in VnDT Treebank.}
\label{table:DPVnDT}
\centering
\begin{tabular}{|c|p{4cm}|r|}
\hline
\multicolumn{1}{|c|}{\textbf{Dependency Type}} & \multicolumn{1}{c|}{\textbf{Description}} & \multicolumn{1}{c|}{\textbf{Percentage}} \\ \hline
                                 NMOD     &Noun modifier          &19.01                   \\ \hline
VMOD     &Verb modifier          &14.81                   \\ \hline
PUNCT     &Punctuation          &13.94                   \\ \hline
*.OB     &Any type ending by OB including DOB (Direct Object), IOB (Indirect Object), and POB (Object of a Preposition)          &11.89                   \\ \hline
SUB     &Subject          &6.80                   \\ \hline
DET     &Determiner          &6.18                   \\ \hline
ADV     &Adverb modifier          &5.92                   \\ \hline
ROOT     &Root          &4.66                   \\ \hline
DEP     &Unclassified          &3.13                   \\ \hline
AMOD     &Adjective modifier          &2.35                   \\ \hline
COORD     &Coordination          &1.88                   \\ \hline
CONJ     &Conjunction          &1.86                   \\ \hline
X.*     &Any dependency type starting with X          &0.28                   \\ \hline
PMOD     &Prepositional modifier          &0.24                   \\ \hline
O.F.Tags     &O.F.Tags refers to other grammatical function tags as dependency types. There are LOC (Location), TMP (Temporal), PRP (Purpose), MNR (Manner), PRD (Predicate),  etc.         &7.05                   \\ \hline
\end{tabular}
\end{table}

\subsection{Dependency Parsing for the Vietnamese Language}
There are only a few studies on dependency parsing for the Vietnamese language. 
The works \cite{DP_Vi_2, DP_Vi_1} built a dependency treebank for the Vietnamese language by converting the Vietnamese Treebank from phrase structures to dependency structures.
The highest accuracies of MSTParser and MaltParser trained on this treebank are 71.66\% (LAS) and 70.49\% (LAS) respectively \cite{DP_Vi_2}. 
We can see that the parsing results reported are still much lower than the accuracies of most of languages in Table \ref{table:performance}. 
This is a challenge for Vietnamese dependency parsing to achieve higher performance in future.
However, the authors did not report a detailed error analysis for further studies.

\section{Error-analysis method}

We characterize parsing errors by the linguistic and structural properties of the dependency graph. The results of our analysis experiments are reported in accuracy, precision, and recall following the labeled scoring scheme. The process of analyzing parsing errors has 2 steps:

\begin{itemize}
\item Step 1: MSTParser and MaltParser are trained and evaluated using the n-fold cross validation scheme.
\item Step 2:  Error analysis is performed based on three types of factors: length factors (sentence length, dependency length), graph factors (distance to root, number of modifier siblings, and non-projective arc degree), and linguistic factors (dependent part-of-speech and dependency type), in a way similar to \cite{Error_2, Error_1}.
\end{itemize}

\begin{figure}[h]
\selectlanguage{english}
\centering
\includegraphics[scale=0.5]{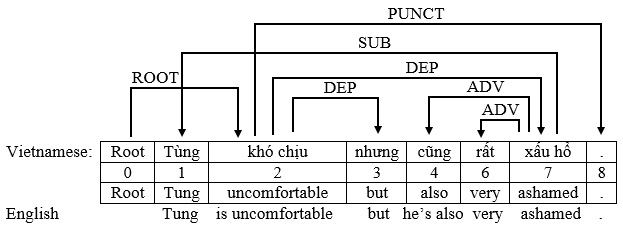}
\caption{A non-projective sentence.}
\label{fig:sentence_2}
\end{figure}

The followings describe the factors in detail:

\begin{itemize}
\item \textbf{Length of a sentence}: The number of words in sentence. The sentence in Figure \ref{fig:sentence_1}  has a length of 7.
\item \textbf{Length of a dependency}: The distance in arc between the head and the dependent. The length of a dependency relation from word $w_i$ to word $w_j$ is equal to \textit{|i – j|}. In Figure \ref{fig:sentence_1}, the dependency arc from \textit{mô tả (describe)} to \textit{kịch bản (scripts)} has a dependency length of 2.
\item \textbf{Distance to root}: For a given arc, this is the number of arcs in the reverse path from the dependent of the arc to the artificial root. For instance, the dependency arc from \textit{Root} to \textit{mô tả (describe)} in Figure \ref{fig:sentence_1} has a distance to root of 2.
\item \textbf{Siblings}: Two dependency arcs \textit{(i, j, l)} and \textit{(i’, j’, l’)} are considered as siblings if they represent modifiers of the same head, i.e, \textit{i = i’}. In Figure \ref{fig:sentence_1}, the arcs from the word \textit{kịch bản (scripts)} to the words \textit{hai (two)} and \textit{mới (new)} are considered as siblings under this definition.
\item \textbf{Non-projective arc degree}: The degree of a dependency arc from word \textit{w} to word \textit{u} is defined as the number of words occurring between w and u that are not descendants of w and modify a word that does not occur between w and u \cite{Nirve_2006}. In the example in Figure \ref{fig:sentence_2}, the arc from \textit{xấu hổ (ashamed)} to \textit{Tùng (Tung)} has a degree of 1.
\item\textbf{Part-of-speech}: Part-of-speech of the dependent. We focus on analyzing the major parts-of-speech such as noun, verb, adjective, adjunct, preposition, and conjunction.
\item \textbf{Dependency type}: Label of dependency arcs. We focus on popular dependency types such as root, subject, object (including direct object and indirect object), noun modifier, verb modifier, adjective modifier, coordination, and conjunction.
\end{itemize}
\footnotetext[4]{MaltOptimizer chooses stacklazy parsing algorithm and SVM learning.}

\section{Parsing Results}

We evaluated the parsers using 5-fold cross validation scheme with the average fold size of 2400 sentences (about 43.750 words). We used the built-in feature set of the two parsers that includes word features, part of speech features, and dependency type features. We did not add any special feature for Vietnamese parsing. 
For MaltParser, we used the MaltOptimizer\footnotemark[4] \cite{MaltOptimizer} in order to choose suitable feature model and parameters and the best parsing algorithm for non-projective structures.  
For MSTParser, we employed the non-projective parsing algorithm. 
In these experiments, we used gold standard part-of-speech tags. 

In general, the accuracy of MSTParser surpassed that of MaltParser. For UAS score, the MSTParser obtained an accuracy of 76.58\% which is 0.5\% higher than that of MaltParser (76.08\%). For LAS score, MSTParser performed better with 70.10\% accuracy, while MaltParser achieved a score of 69.88\%. It should be noted that we included punctuation dependencies in calculation of the scores. 

\section{Error Analysis Results}

The analysis results are measured on the test and parsed (predicted) data. We compare our results with those reported in the paper \cite{Error_1} in several aspects. Our error anlysis results are presented in detail below.

\subsection{Length Factors}
Table \ref{table:SentenceLength} shows the accuracy of the two parsers relative to sentence length. Difference between the two parsing accuracies ($\Delta$) is less than 1.0. In general, both of the parsers tend to have higher accuracies for shorter sentences. Similar to other languages, MaltParser tends to perform better on short sentences (sentence length of 11--20 accounts for 35.39\%). 
This is because the greedy inference algorithm employed by MaltParser makes fewer parsing decisions. As a result, the likelihood of error propagation is reduced when parsing short sentences. The rich feature representation also increases the performance of MaltParser.

\begin{table}[h]
\selectlanguage{english}
\centering
\caption{Accuracy relative to sentence length.}
\label{table:SentenceLength}
\begin{tabular}{|c|c|r|r|r|}
\hline
Sentence Length & \multicolumn{1}{c|}{\textbf{Percentage}} & \multicolumn{1}{c|}{\textbf{MST}} & \multicolumn{1}{c|}{\textbf{Malt}} & \multicolumn{1}{c|}{\textbf{$\Delta$}} \\ \hline
1--10           & 19.00                               & 79.73                    & 79.47                     & 0.26                       \\ \hline
11--20          & 35.39                               & 73.23                    & 73.35                     & -0.12                      \\ \hline
21--30          & 25.61                               & 70.02                    & 69.93                     & 0.09                       \\ \hline
31--40          & 12.17                               & 67.43                    & 67.33                     & 0.10                       \\ \hline
41-50           & 4.94                                & 65.77                    & 65.94                     & -0.17                      \\ \hline
\textgreater50  & 2.89                                & 64.74                    & 63.89                     & 0.85                       \\ \hline
\end{tabular}
\end{table}

In Vietnamese, short dependencies are often noun modifiers, prepositional objects, adjective modifiers and direct objects. Long sentences contain increasingly complex syntactic structures, resulting in long dependencies which often contain prepositions, conjunctions, or multiple clauses (see the column of average dependency lengths in Table \ref{table:DependencyType}). These characteristics affect the accuracy of the parser.

Table \ref{table:DependencyLength} measures the precision and recall of each parser relative to dependency length in predicted and gold standard dependency graphs. 
Precision measures the percentage of predicted arcs of length d that were correct. Recall presents the percentage of gold standard arcs of length d that were predicted correctly. We can see that both parsers tend to have higher precision for shorter dependency lengths. MSTParser is more precise than MatlParser for shorter dependency length (length of 1 and 2 accounts for 67.97\%), especially dependency length of 1 with $\Delta$ = 1.15.  MaltParser is far more precise for longer dependency arcs (dependency length >2) except dependency length of 12. 
For the recall measure, MSTParser is better than MaltParser for longer dependency lengths. 
These observations are contrary to those of the analysis of the other languages. 

Theoretically, MSTParser should not perform better or worse for arcs of any length \cite{Error_1}. However, it can be seen that the number of dependency arcs with a length larger than 2 is smaller than the number with length less than or equal to 2. 
The rich feature space employed by the MaltParser can efficiently reduced the phenomenon of error propagation, which could be the reason why the overall accuracies of the two parsers are nearly identical. 

\begin{table}[h]
\selectlanguage{english}
\centering
\caption{Dependency arc precision/recall relative to predicted/gold dependency length.}
\label{table:DependencyLength}
\begin{tabular}{|c|c|r|r|r|r|r|r|}
\hline
\multirow{2}{*}{\textbf{Length}} & \multicolumn{1}{c|}{\multirow{2}{*}{\textbf{Percentage}}} & \multicolumn{3}{c|}{\textbf{Precision}}                                                                       & \multicolumn{3}{c|}{\textbf{Recall}}                                                                         \\ \cline{3-8} 
                             & \multicolumn{1}{c|}{}                               & \multicolumn{1}{c|}{\textbf{MST}} & \multicolumn{1}{c|}{\textbf{Malt}} & \multicolumn{1}{c|}{\textbf{$\Delta$}} & \multicolumn{1}{c|}{\textbf{MST}} & \multicolumn{1}{c|}{\textbf{Malt}} & \multicolumn{1}{c|}{\textbf{$\Delta$}} \\ \hline
1&    50.18       &85.13    &83.98     &1.15      &81.91      &82.07     &-0.16       \\ \hline
2&    17.79       &67.01    &66.75     &0.26      &65.38      &66.56     &-1.18       \\ \hline
3&    8.14       &56.20    &56.66     &-0.46      &56.78      &58.81     &-2.03       \\ \hline
4&    4.75       &49.53    &50.05     &-0.52      &51.73      &52.71     &-0.98       \\ \hline
5&    3.37       &47.05    &48.17     &-1.12      &49.76      &50.14     &-0.38       \\ \hline
6&    2.46       &42.69    &44.86     &-2.17      &48.16      &47.83     &0.33       \\ \hline
7&    1.86       &43.97    &45.45     &-1.48      &47.61      &46.23     &1.38       \\ \hline
8&    1.53       &42.94    &43.52     &-0.58      &46.82      &45.52     &1.30       \\ \hline
9&    1.25       &40.34    &43.06     &-2.72      &46.17      &45.07     &1.10       \\ \hline
10&    1.05       &38.46    &42.13     &-3.67      &42.57      &43.84     &-1.27       \\ \hline
11&    0.90       &42.69    &43.24     &-0.55      &45.45      &45.04     &0.41       \\ \hline
12&    0.77       &44.09    &42.96     &1.13      &47.66      &44.53     &3.13       \\ \hline
13&    0.69       &45.35    &48.20     &-2.85      &50.91      &49.76     &1.15       \\ \hline
14&    0.57       &42.62    &45.62     &-3.00      &47.29      &46.94     &0.35       \\ \hline
15&    0.50       &45.59    &47.86     &-2.27      &52.25      &47.49     &4.76       \\ \hline
$\ge$16&    4.19       &47.72    &53.34     &-5.62      &59.75      &54.66     &5.09       \\ \hline
\end{tabular}
\end{table}

\subsection{Graph Factors}

Table \ref{table:DistanceToRoot} shows the precisions and recalls of dependency arcs relative to distance to root. Precision is the percentage of dependency arcs in the predicted graphs at a distance of d that is correct. Recall is the percentage of dependency arcs in the gold standard graphs at a distance of d that is predicted.  As a result, both parsers have low precision and recall, and tend to be more precise for distances of 2 and 3. The figures in this table shows that for arcs close to the root, MSTParser is better than MaltParser, and the reverse trend is observed for arcs further away from the root. 
For MaltParser, dependency arcs further away from the root are usually constructed early by the parsing algorithm. Words that are not assigned as modifiers are automatically linked to the root. Therefore, MaltParser has a low precision for root modifiers, and these results are consistent with their previous analysis results. 

\begin{table}[h]
\selectlanguage{english}
\caption{Dependency arc precision/recall relative to predicted/gold distance to root.}
\label{table:DistanceToRoot}
\centering
\begin{tabular}{|c|c|r|r|r|r|r|r|}
\hline
\multirow{2}{*}{\textbf{Distance}} & \multicolumn{1}{c|}{\multirow{2}{*}{\textbf{Percentage}}} & \multicolumn{3}{c|}{\textbf{Precision}}                                                                      & \multicolumn{3}{c|}{\textbf{Recall}}                                                                         \\ \cline{3-8} 
                                           & \multicolumn{1}{c|}{}                               & \multicolumn{1}{c|}{\textbf{MST}} & \multicolumn{1}{c|}{\textbf{Malt}} & \multicolumn{1}{c|}{\textbf{$\Delta$}} & \multicolumn{1}{c|}{\textbf{MST}} & \multicolumn{1}{c|}{\textbf{Malt}} & \multicolumn{1}{c|}{\textbf{$\Delta$}} \\ \hline
 1      &15.05      &59.93     &46.62      &13.31      &60.43      &61.55     &-1.12      \\ \hline
2      &37.22      &61.92     &64.05      &-2.13      &61.29      &65.58     &-4.29      \\ \hline
3      &34.31      &61.41     &65.46      &-4.05      &59.05      &61.01     &-1.96      \\ \hline
4      &9.48      &44.56     &56.66      &-12.10      &45.54      &48.86     &-3.32      \\ \hline
5      &2.81      &31.98     &48.25      &-16.27      &36.98      &38.76     &-1.78      \\ \hline
6      &0.88      &20.27     &49.50      &-29.23      &27.19      &37.01     &-9.82      \\ \hline
$\ge$7      &0.24      &11.97     &24.55      &-12.58      &31.36      &29.42     &1.94      \\ \hline
\end{tabular}
\end{table}

The second graph property we examine is the sibling of arcs. We want to quantify the local neighborhood of an arc within a dependency graph. Table \ref{table:NumberOfModifierSiblings} measures the precisions and recalls of the parsers relative to the number of predicted and gold-standard siblings of dependency arcs. In general, both of parsers have low precisions and recalls, and tend to be more precise for those arcs with fewer siblings. 
MSTParser performs better for dependency arcs that have no siblings (dependency relations containing dependent is leaf), whereas MaltParser tends to be more accurate for those arcs with more siblings. 
When a sentence has more siblings, it generates more long-distance dependencies. 
These results are consistent with the previous analysis results. 

Table \ref{table:Degree} shows the precisions and recalls relative to different arc degrees in predicted and gold standard non-projective dependency graphs.  
MSTParser recognizes more degrees than MaltParser. 
In general, both of the parsers yield low precisions and recalls. 
These results can be explained by two reasons.
First, the training and test data sets contain too few non-projective dependency structures (from 3\% to 5\%). 
Second, a high degree of non-projectivity corresponds to longer dependencies which are more challenging for the parsers to predict correctly. 

\begin{table}[h]
\selectlanguage{english}
\caption{Dependency arc precisions/recalls relative to predicted/gold siblings. NMS = Number of Modifier Siblings; PER = Percentage.}
\label{table:NumberOfModifierSiblings}
\centering
\begin{tabular}{|c|r|r|r|r|r|r|r|}
\hline
\multirow{2}{*}{\textbf{NMS}} & \multicolumn{1}{c|}{\multirow{2}{*}{\textbf{PER}}} & \multicolumn{3}{c|}{\textbf{Precision}}                                                                      & \multicolumn{3}{c|}{\textbf{Recall}}                                                                         \\ \cline{3-8} 
                                                      & \multicolumn{1}{c|}{}                               & \multicolumn{1}{c|}{\textbf{MST}} & \multicolumn{1}{c|}{\textbf{Malt}} & \multicolumn{1}{c|}{\textbf{$\Delta$}} & \multicolumn{1}{c|}{\textbf{MST}} & \multicolumn{1}{c|}{\textbf{Malt}} & \multicolumn{1}{c|}{\textbf{$\Delta$}} \\ \hline
0    &35.77      &80.83      &78.66        &2.17         &77.04         &77.37      &-0.33       \\ \hline
1    &35.77      &44.92      &49.19        &-4.27         &47.04         &50.28      &-3.24       \\ \hline
2    &15.04      &28.50      &36.57        &-8.07         &33.10         &38.57      &-5.47       \\ \hline
3    &5.65      &25.52      &32.50        &-6.98         &28.92         &33.63      &-4.71       \\ \hline
4    &3.09      &22.46      &28.80        &-6.34         &25.29         &29.02      &-3.73       \\ \hline
5    &1.73      &25.16      &31.97        &-6.81         &24.70         &31.46      &-6.76       \\ \hline
6    &1.16      &24.26      &28.64        &-4.38         &20.37         &26.86      &-6.49       \\ \hline
7    &0.76      &22.01      &25.01        &-3.00         &16.33         &23.80      &-7.47       \\ \hline
8    &0.45      &20.80      &22.20        &-1.40         &10.97         &19.86      &-8.89       \\ \hline
9    &0.27      &21.42      &17.53        &3.89         &8.95         &13.49      &-4.54       \\ \hline
$\ge$10    &0.36      &9.04      &9.14        &-0.10         &2.96         &11.18      &-8.22       \\ \hline
\end{tabular}
\end{table}
\begin{table}[h]
\selectlanguage{english}
\centering
\caption{Dependency arc precision/recall relative to predicted/gold degree of non-projectivity.}
\label{table:Degree}
\begin{tabular}{|c|r|r|r|r|}
\hline
\multirow{2}{*}{\textbf{Degree}} & \multicolumn{2}{c|}{\textbf{MST}}                                              & \multicolumn{2}{c|}{\textbf{Malt}}                                             \\ \cline{2-5} 
                                 & \multicolumn{1}{c|}{\textbf{Precision}} & \multicolumn{1}{c|}{\textbf{Recall}} & \multicolumn{1}{c|}{\textbf{Precision}} & \multicolumn{1}{c|}{\textbf{Recall}} \\ \hline
0            &5.21             &51.33               &8.99               &2.50              \\ \hline
1            &0.06             &0.91               &0.00               &0.00              \\ \hline
2            &0.25             &0.95               &N/A               &N/A              \\ \hline
3            &0.67             &5.00               &N/A               &N/A              \\ \hline
\end{tabular}
\end{table}

\subsection{Linguistic Factors}


Table \ref{table:DependentPOS} shows the accuracies of the two parsers for different parts of speech. 
We measure labeled dependency accuracy relative to the part-of-speech of the modifier word in dependency relations. 
As a result, we can see that both of the parsers achieve very high performances for adjuncts, quantities, and determiners ($\ge$91.00\%). 
For dependent parts-of-speech corresponding to lower accuracies, we see that MaltParser performs slightly better for verbs, conjunctions and punctuations (comma, period, quotation mark), while MSTParser performs better on the other categories including nouns, prepositions, adjectives, pronouns, and particles. 
The difference between the accuracies of two parsers ($\Delta$) varies from 0.1\% to 4.5\%.

Table \ref{table:DependentDP} shows the dependent parts-of-speech and possible dependency types for each POS. 
We can see the correlation between the accuracies of dependent part-of-speech and its dependency types. For instance, MaltParser achieves high accuracies for verbs and conjunctions related to the types which  appear with high frequencies: VMOD, NMOD, DEP, and COORD. 
On the other hand, MSTParser is better on categories such as nouns, pronouns, prepositions, and adjectives. Because these categories are related to high-frequency dependency relations (SUB, DET, POB, DOB, LOC, and DET). 

\begin{table}[t!]
\selectlanguage{english}
\centering
\caption{Dependency arc precision/recall relative to predicted/gold degree of non-projectivity.}
\label{table:DependentPOS}
\begin{tabular}{|c|r|r|r|}
\hline
\multicolumn{1}{|c|}{\textbf{Dependent POS}} &\multicolumn{1}{|c|}{\textbf{MST}} &\multicolumn{1}{|c|}{\textbf{Malt}} &\multicolumn{1}{|c|}{\textbf{$\Delta$}}\\ \hline
Noun &73.87 &72.31 &1.56\\ \hline
Verb &63.33 &64.13 &-0.80 \\ \hline
Adjunct &92.85 &92.55 &0.30 \\ \hline
Preposition &52.91 &51.03 &1.88 \\ \hline
Adjective &69.21 &68.19 &1.02 \\ \hline
Pronoun &79.61 &79.21 &0.40 \\ \hline
Conjunction &47.25 &51.28 &-4.03 \\ \hline
Determiner &98.76 &98.88 &-0.12 \\ \hline
Quantity &91.76 &91.12 &0.64 \\ \hline
Particle &72.06 &70.17 &1.89 \\ \hline
Punctuation &62.93 &65.65 &-2.72 \\ \hline
\end{tabular}
\end{table}

\begin{table}[b!]
\selectlanguage{english}
\caption{Precision/recall for different dependency types. DT = dependency type; Per = percentage; DLA = dependency length average.}
\label{table:DependencyType}
\centering
\begin{tabular}{|c|c|c|r|r|r|r|}
\hline
\multirow{2}{*}{\textbf{DT}} & \multirow{2}{*}{\textbf{PER}} & \multirow{2}{*}{\textbf{DLA}} & \multicolumn{2}{c|}{\textbf{Precision}} & \multicolumn{2}{c|}{\textbf{Recall}} \\ \cline{4-7} 
                    &                       &                      & \textbf{MST}           & \textbf{Malt}           &\textbf{ MST}          & \textbf{Malt}         \\ \hline
NMOD       &19.49        &1.83              &78.22      &79.04      &77.07      &75.09                    \\ \hline
VMOD      &14.77        &2.58              &59.48      &60.70      &58.17      &58.22                    \\ \hline
PUNCT       &14.17        &8.39              &61.32      &64.00      &61.32      &64.34                    \\ \hline
SUB       &6.76        &3.57              &66.47      &65.70      &68.66      &67.29                    \\ \hline
DET       &6.28        &1.21              &94.54      &93.94      &93.49      &93.77                    \\ \hline
DOB       &5.89        &1.63              &76.19      &68.76      &64.38      &64.00                   \\ \hline
ADV       &5.83        &1.45              &92.80      &92.72      &93.69      &93.61                    \\ \hline
POB       &5.56        &1.27              &96.75      &95.87      &93.81      &93.22                    \\ \hline
ROOT       &4.69        &5.62              &82.69      &79.92      &82.69      &79.84                    \\ \hline
DEP       &3.13        &7.29              &33.47      &41.51      &51.18      &47.83                    \\ \hline
AMOD       &2.25        &1.50              &73.51      &72.01      &71.71      &69.12                    \\ \hline
LOC       &2.28        &2.59              &52.21      &45.93      &49.54      &50.18                    \\ \hline
TMP       &2.14        &5.40              &42.29      &38.99      &46.46      &50.62                    \\ \hline
COORD       &1.88        &5.64              &48.52      &50.94      &43.68      &50.74                    \\ \hline
CONJ       &1.87        &2.43              &75.44      &70.32      &67.01      &69.85                    \\ \hline
PRP       &1.28        &3.98              &32.26      &32.47      &45.61      &42.28                    \\ \hline
MNR       &0.39        &3.84              &25.96      &21.44      &40.71      &40.04                    \\ \hline
PRD       &0.32        &5.83              &9.25      &1.00      &16.41      &15.78                    \\ \hline
PMOD       &0.24        &4.81              &42.92      &41.03      &46.71      &40.81                    \\ \hline
IOB       &0.20        &2.80              &25.69      &27.45      &36.69      &37.95                    \\ \hline
\end{tabular}
\end{table}

Table \ref{table:DependencyType} displays precisions and recalls for different dependency types. In addtion, this table also gives us some information such as the percentage of dependency types and the average of dependency lengths. MSTParser tends to be more accurate for shorter dependency-lengths with the relations such as SUB, DET, DOB, ADV, PMOD, AMOD, LOC, and CONJ, while MaltParser has higher precision for longer dependency-lengths with the relations such as PUNCT, DEP, COORD, and PRD. 
These results is consistent with the previous results in Table \ref{table:DependencyLength}. 
However, we can also observe that some cases reverse the general trend, i.e, NMOD, VMOD, and ROOT.
In particular, dependency relations achieving high accuracies are DET, POB and ADV.  Because they have short dependency lengths (LDA < 1.5) and fixed-word heads. 
And dependency relation ADV primarily links verb to adverb and appears frequently in training set. 

We consider precision and recall for dependents of the root node (mostly verbal predicate, noun predicate, and adjective predicate), and for subjects and objects (direct objects and indirect objects). 
MSTParser has considerably better precision (and better recall) for the root, subject, and direct object relations, but MaltParser is better for indirect objects. The accuracy of verbal predicate (ROOT-Verb) is more precise for MSTParser, shown in Table \ref{table:RootRelations}. As results of root distance, we seen that MSTParser is precise for root modifiers.

\begin{table}[h]
\selectlanguage{english}
\centering
\caption{Accuracy of root relation.}
\label{table:RootRelations}
\begin{tabular}{|l|r|r|r|}
\hline
\multirow{2}{*}{\textbf{Root relation}} & \multirow{2}{*}{\textbf{Percentage}} & \multicolumn{2}{c|}{\textbf{Accuracy}} \\ \cline{3-4} 
                          &                       & \textbf{MST}           &\textbf{ Malt}         \\ \hline
Root - Verb               & 88.80                 & 86.46         & 82.98         \\ \hline
Root - Noun               & 5.80                  & 56.82         & 68.77         \\ \hline
Root - Adjective          & 4.70                  & 50.92         & 45.71         \\ \hline
\end{tabular}
\end{table}

In term of COORD and CONJ, MSTParser is better for CONJ, but MaltParser has better precision for COORD. We can see that COORD often links a noun to a conjunction with long dependency length (5.64). Linking a conjunction to a noun forms a relation as CONJ with short dependency length (2.43). Accuracies of relations related to conjunctions are also consistent with previous results.

All experiments in this section show that there is a trade-off between global learning and inference in the MSTParser and rich feature presentation in the MaltParser.  
Although the accuracy of MaltParser is affected by error propagation,
the rich feature presentation employed by the model helps to overcome the problem in many cases. 
As a result, it performs rather well for long dependencies, whereas MSTParser is more precise for short dependencies. 
This is the main difference between the Vietnamese language and other languages (from average of 13 languages in paper \cite{Error_1}) .

\begin{table}[h!]
\selectlanguage{english}
\caption{Dependent POS and a list of its dependency types.}
\label{table:DependentDP}
\centering
\begin{tabular}{|p{2cm}|p{6cm}|}
\hline
\multicolumn{1}{|c|}{\textbf{Dependent POS}} & \multicolumn{1}{c|}{\textbf{Dependency Type}} \\ \hline
Adjunct      &adv (76.45\%), amod (12.88\%), nmod (7.90\%), dep (2.50\%), others(0.27\%).                    \\ \hline
Determiner      &det (99.59\%), others(0.41\%).                    \\ \hline
Quantity      &det (95,32\%), nmod (1.45\%), others(3.23\%).                    \\ \hline
Verb      &vmod (50.62\%), root (21\%), nmod (9.53\%), conj (7.19\%), dep (5.80\%), others(5.86\%).                    \\ \hline
Conjunction      &coord (50.48\%), dep (32.39\%), nmod (8.34\%), vmod (5.73\%), amod (2.96\%), others(0.37\%).                    \\ \hline
Noun      &nmod (33.17\%), dob (16.97\%), sub (16.69\%), pob (15.64\%), others(17.53\%).                    \\ \hline
Pronoun      &sub (31.61\%), det (21.92\%), nmod (17.69\%), pob (11.56\%), dob (5.89\%), vmod (4.17\%), tmp (2.59\%), others(4.57\%).                    \\ \hline
Preposition      &loc (23.49\%), nmod (22.30\%), vmod (15.34\%), prp (10.85\%), pmod (3.58\%), mnr (3.21\%), dir (3.08\%), iob (3.00\%), amod (2.28\%), others(12.87\%).                    \\ \hline
Adjective      &nmod (49.77\%), vmod (24.35\%), amod (5.75\%), dep (4.69\%), root (4.52\%), prd (3.27\%), others(7.65\%).                    \\ \hline
Particle      &nmod (36.34\%), vmod (29.39\%), amod (13.98\%), dep (13.69\%), det (6.52\%), others(0.08\%).                    \\ \hline
Punctuation      &punct (100\%).                    \\ \hline
\end{tabular}
\end{table}

\subsection{Comparison with Other Languages}


Each language has its own characteristics which lead to differences in the accuracy relative to the part-of-speech tags as well as the dependency types. 
Table \ref{table:comparing_1} and Table \ref{table:comparing_2} show that the difference between the parsing accuracies of the Vietnamese language and the average accuracies of 13 languages (listed in Table \ref{table:performance}) relative to the dependency type is significant.
The average accuracies of 13 languages in these tables are taken from paper \cite{Error_2}.

The parsing results for adverbs for Vietnamese are higher in comparison with the other languages.
However for other types of parts-of-speech, the accuracies of the Vietnamese parsing are considerably lower, especially verbs, adjectives, and conjunctions.
Roots, subjects and objects in Vietnamese are far less precise than in others.  One reason for such low accuracy could be that the VnDT treebank contains many long sentences with complex structures (see the statistics in Section II.B). 
Inconsistencies in the Vietnamese Treebank may also affect as presented in paper \cite{Quy_2013}. 
Other reasons may come from the difference between the characteristics of Vietnamese and others, which requires further effort of the research community.

\footnotetext[5]{In Vietnamese, an adjunct is often an adverb which used to modify a verb.}
\begin{table}[h]
\selectlanguage{english}
\caption{Accuracy relative to dependent part-of-spech of Vietnamese (ViL) versus average of 13 languages (Others).}
\label{table:comparing_1}
\centering
\begin{tabular}{|l|r|r|r|r|r|r|}
\hline
\multicolumn{1}{|c|}{\multirow{2}{*}{\textbf{Dependent POS}}} & \multicolumn{3}{c|}{\textbf{MST}}                                                        & \multicolumn{3}{c|}{\textbf{Malt}}                                                       \\ \cline{2-7} 
\multicolumn{1}{|c|}{}                    & \multicolumn{1}{c|}{\textbf{ViL}} & \multicolumn{1}{c|}{\textbf{Others}} & \multicolumn{1}{c|}{\textbf{$\Delta$}} & \multicolumn{1}{c|}{\textbf{ViL}} & \multicolumn{1}{c|}{\textbf{Others}} & \multicolumn{1}{c|}{{$\Delta$}} \\ \hline
Verb         &63.3      &82.6       &-19.3        &63.6         &81.9        &-18.3             \\ \hline
Noun         &73.9      &80.0       &-6.1        &72.3         &80.7        &-8.4             \\ \hline
Pronoun         &79.6      &88.4       &-8.8        &79.5         &89.2        &-9.7             \\ \hline
Adjective         &69.2      &89.1       &-19.9        &68.1         &87.9        &-19.8             \\ \hline
Adverb\footnotemark[5]         &92.9      &78.3       &14.6        &92.6         &77.4        &15.2             \\ \hline
Conjunction         &47.3      &73.1       &-25.8        &48.8         &69.8        &-21.0             \\ \hline
\end{tabular}
\end{table}


\begin{table}[h]
\selectlanguage{english}
\caption{Precision relative to dependency type of Vietnamese (ViL) versus average of 13 languages (Others).}
\label{table:comparing_2}
\centering
\begin{tabular}{|l|r|r|r|r|r|r|}
\hline
\multicolumn{1}{|c|}{\multirow{2}{*}{\textbf{Dependency Type}}} & \multicolumn{3}{c|}{\textbf{MST}}                                                        & \multicolumn{3}{c|}{\textbf{Malt}}                                                       \\ \cline{2-7} 
\multicolumn{1}{|c|}{}                    & \multicolumn{1}{c|}{\textbf{ViL}} & \multicolumn{1}{c|}{\textbf{Others}} & \multicolumn{1}{c|}{\textbf{$\Delta$}} & \multicolumn{1}{c|}{\textbf{ViL}} & \multicolumn{1}{c|}{\textbf{Others}} & \multicolumn{1}{c|}{\textbf{$\Delta$}} \\ \hline
Root         &82.7      &89.9       &-7.2        &79.0         &84.7        &-5.7             \\ \hline
Subject         &66.5      &79.9       &-13.4        &66.5         &80.3        &-13.8             \\ \hline
Object         &50.9      &76.5       &-25.6        &49.7         &77.2        &-27.5             \\ \hline
\end{tabular}
\end{table}

\section{Conclusion and future directions}
In this paper, we have presented a thorough study of distinctive error distributions produced by MSTParser and MaltParser for the Vietnamese language.
In particular, we provide a comparison with the error analysis of dependency parsing for otherss. 
This would be helpful for researchers to create better parsing models. 

Based on the analysis results, we suggest some possible directions for the improvement of data-driven dependency parsing for the Vietnamese language in future:

\begin{enumerate}
\item We can study a way to represent feature models suitable for Vietnamese language. For instance, we can focus on improving the subject (SUB) dependency that links  the main verb to its modifiers. 
This work leads to the improvement of POS features relative to subjects such as nouns and verbs. 

\item Vietnamese dependency parsing can be improved by integrating the strength of both graph-based and transition-based models in a similar way as proposed by Nirve and McDonald \cite{Integrating}.

\item We can also build ensemble systems as proposed by Sagae and Lavie (2006) \cite{Sagae}. The error analysis for a range of linguistic and graph-based factors in this paper can help to build the weighing schemes for ensemble systems.
\end{enumerate}

\selectlanguage{english}

\end{document}